\documentclass{article}

\usepackage{arxiv}
\usepackage[utf8]{inputenc} % allow utf-8 input
\usepackage[T1]{fontenc}    % use 8-bit T1 fonts
\usepackage{hyperref}       % hyperlinks
\usepackage{url}            % simple URL typesetting
\usepackage{booktabs}       % professional-quality tables
\usepackage{amsfonts}       % blackboard math symbols
\usepackage{nicefrac}       % compact symbols for 1/2, etc.
\usepackage{microtype}      % microtypography
\usepackage{lipsum}
\usepackage{graphicx}
\usepackage{subfigure}
\usepackage{svg}
\usepackage{amsmath}
\usepackage{geometry}
\usepackage{tabularx}
\usepackage{adjustbox}
\usepackage{rotating}
\graphicspath{ {./images/} }

\title{Bridging the Semantic Gaps: Improving MVQA Consistency with LLM-Augmented Question Sets}

\author{
 Yongpei Ma \\
  School of Computer Science\\
   University of Sydney\\
  Darlington, NSW 2008 \\
  \texttt{yongpei.ma@sydney.edu.au} \\
   \And
 Pengyu Wang \\
  School of Computer Science\\
   University of Sydney\\
  Darlington, NSW 2008 \\
  \texttt{pwan0442@uni.sydney.edu.au} \\
  \And
 Adam Dunn \\
  School of Medical Sciences\\
  University of Sydney\\
  Darlington, NSW 2008 \\
  \texttt{adam.dunn@sydney.edu.au} \\
    \And
 Usman Naseem \\
  School of Coumputing\\
  Macquarie University\\
  Macquarie Park, NSW 2113 \\
  \texttt{usman.naseem@mg.edu.au} \\
    \And
 Jinman Kim \\
  School of Computer Science\\
   University of Sydney\\
  Darlington, NSW 2008 \\
  \texttt{jinman.kim@sydney.edu.au} \\
}

\begin{document}
\maketitle
\begin{abstract}

Medical Visual Question Answering (MVQA) systems hold promise for assisting clinicians by interpreting medical images in response to natural language queries. However, linguistic variability in question phrasing often undermines the consistency of these systems. To address this challenge, we propose a Semantically Equivalent Question Augmentation (SEQA) framework, which leverages large language models (LLMs) to generate diverse yet semantically equivalent rephrasings of questions. Specifically, this approach enriches linguistic diversity while preserving semantic meaning, and thereby enhancing the robustness of MVQA models. We further introduce a novel evaluation metric, Total Agreement Rate with Semantically Equivalent Input and Correct Answer (TAR-SC), which assesses a model's capability to generate consistent and correct responses to semantically equivalent linguistic variations. In addition, we also propose other three diversity metrics - average number of QA items per image (ANQI), average number of questions per image with the same answer (ANQA), and average number of open ended questions per image with the same semantics (ANQS). Using the SEQA framework, we augmented the widely benchmarked MVQA public datasets of SLAKE, VQA-RAD, and PathVQA. As a result, all three datasets achieved significant improvements by incorporating more semantically equivalent questions: ANQI increased by an average of 86.1, ANQA by 85.1, and ANQS by 46. Subsequent experiments evaluate three MVQA models (M2I2, MUMC, and BiomedGPT) under both zero-shot and fine-tuning settings on the enhanced datasets. Experimental results in MVQA benchmark datasets demonstrate that fine-tuned models achieve an average accuracy improvement of 19.35\%, while our proposed TAR-SC metric shows an average improvement of 11. 61\%, indicating a substantial enhancement in model consistency. These results highlight the importance of exposing models to varied question formulations, ultimately improving their clinical applicability.

\end{abstract}

\begin{figure*}
\centering
\includegraphics[width=1\textwidth]{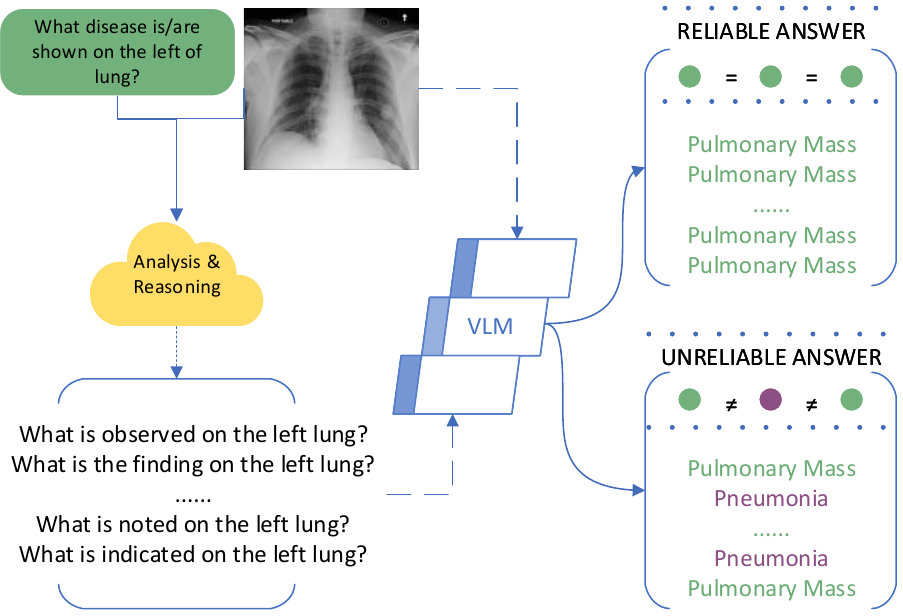}
\caption{Overview of our proposed Semantically Equivalent Question Augmentation (SEQA) framework. Original question and image are used as an input into an LLM, such as Gemini or GPT-4, to generate linguistically diversified  questions. These augmented questions, along with the image, are then fed into a vision-language model (VLM). If the model consistently provides the same answer for all the augmented questions, it is deemed as a consistent model.}
\label{fig_1}
\end{figure*}

% keywords can be removed
%\keywords{First keyword \and Second keyword \and More}
\section{Introduction}
Medical Visual Question Answering (MVQA) can support clinical decision making by providing accurate and timely interpretations of medical images in response to natural language queries \cite{antol2015vqa}. The integration of large language models (LLMs) with MVQA has shown promise in enhancing diagnostic accuracy and facilitating clinical decision-making \cite{singhal2025toward}. LLMs, such as GPT-3 and its successors, have demonstrated capability in natural language understanding and generation, which can be used to interpret and answer complex questions about medical images \cite{brown2020language}.

Despite these advancements, the consistency and reliability of MVQA models remain a critical barrier to their use in clinical practice. Consistency is a model's ability to produce the same or similar outputs when given the same or equivalent inputs, emphasizing its stability and predictability in performance. Reliability is a model's general dependability across a variety of conditions and inputs, which can be measured with accuracy of the models' results. Consistency is closely associated with reliability \cite{khan2024consistency}. Existing MVQA datasets, such as VQA-RAD \cite{lau2018dataset}, SLAKE \cite{liu2021slake}, and PathVQA \cite{he2020pathvqa}, are used extensively for developing and evaluating MVQA models. However, these datasets often lack sufficient linguistic diversity in the phrasing of questions \cite{lau2018dataset, liu2021slake}. Consequently, models trained on these datasets may struggle to generalize to real-world scenarios where questions may be paraphrased or restructured where variability in question phrasing, synonymous terms, and linguistic nuances, can lead to inconsistent answers from MVQA models, undermining their consistency. For instance, different clinicians may ask semantically identical questions differently, producing different answers. As such, inconsistency is a strong indicator of unreliability, and issues of consistency are inherently issues of model reliability. The implications of measuring consistency in VQA model is paramount in clinical applications, where inconsistent or incorrect information can lead to unintended consequences \cite{esteva2019guide}.

To address these challenges, we propose a novel approach to enhance the consistency of MVQA models by augmenting existing datasets with semantically equivalent questions generated using LLMs (see Figure \ref{fig_1}). By leveraging the language generation capabilities of LLMs, we create multiple paraphrased versions of each original question, ensuring that the corresponding answers remain the same. Our contributions are as follows:

\begin{itemize}
\item We propose a new MVQA consistency framework named Semantically Equivalent Question Augmentation (SEQA), which uses LLMs to generate paraphrased versions of questions to enrich the datasets and improve language diversity.  Then, we build a consistent VLM using the processed datasets.
\item We provide two augmented MVQA datasets, SLAKE En 1.0-LD and VQA-RAD-LD, where LD stands for \textit{Language Diversity}. These datasets are created by applying the SEQA framework to SLAKE En 1.0 and VQA-RAD, respectively.  
\item We introduce three dataset-specific metrics: Average Number of QA Items per Image (ANQI), Average Number of Questions per Image with the Same Answer (ANQA), and Average Number of Open-Ended Questions per Image with the Same Semantics (ANQS). These metrics are used to assess the richness of semantic expression within the dataset.
\item We also introduce the Total Agreement Rate with Similar Input and Correct Answer (TAR-SC) building on TAR \cite{atil2024llm}. This is a novel metric that measures the consistency of models in providing correct answers to semantically similar questions. 
\item We demonstrate that fine-tuned models on datasets augmented with the SEQA framework improves their consistency in answering semantically equivalent questions. 
\end{itemize}

The augmented datasets and the framework for augmenting VQA datasets will be available on GitHub.

\section{Related Work}
\subsection{Large Language Models for Medical Visual Question Answering}
The integration of LLMs with MVQA systems has emerged as an approach to improving diagnostic accuracy and facilitating clinical decision-making. LLMs, such as GPT-3 and its successors, have demonstrated capabilities in natural language understanding and generation, which can be leveraged in the domain of MVQA to interpret and answer complex questions based on medical images.

Early approaches to MVQA primarily focused on discriminative models, where the task was framed as a classification problem. These models, such as those used in the PathVQA dataset, aimed to predict answers from a predefined set of options \cite{he2020pathvqa}. While effective in closed-set scenarios, these models were limited by their inability to generate novel answers or handle open-ended questions, a critical requirement in real-world medical applications.

To address these limitations, recent research has explored the use of generative models, which can produce free-form text answers. For example, the work by Nguyen et al.  \cite{ben2021overview} and the VQA-Med framework \cite{ben2019vqa}, utilizes LLMs to generate answers based on the visual features extracted from medical images. These models have shown improved performance in handling open-ended questions, providing more flexibility and applicability in clinical settings.

The incorporation of VLMs, such as those described by Zhang et al.  \cite{zhang2023large}, has further enhanced the ability of these systems to understand and reason about complex visual and textual information. These models combine the strengths of LLMs with advanced computer vision techniques to interpret medical images and generate accurate and contextually relevant answers.

Another significant advancement is the development of self-supervised pretraining techniques for LLMs, which have been shown to improve the performance of MVQA systems. For instance, the work by Chen et al.  \cite{chen2020generative} demonstrated that self-supervised vision-language pretraining could significantly enhance the model’s ability to generalize to new medical images and questions, reducing the reliance on large, labeled datasets.

In addition to these innovations, researchers have also focused on improving the interpretability and explainability of LLM-based MVQA systems. Models such as the Explainable AI (XAI) VQA framework by Ali et al.  \cite{ALI2023101805} aim to provide transparent and understandable explanations for the answers generated by the system, which is crucial for gaining the trust of clinicians and ensuring the safe deployment of these technologies in healthcare.

Despite the progress made, challenges remain in developing robust MVQA systems. One of the key issue is the lack of large, high-quality annotated datasets that are representative of the diversity and complexity of real-world medical scenarios. Efforts are being made to address this gap, with new datasets such as the SLAKE dataset \cite{liu2021slake} and the VQA-RAD dataset \cite{lau2018dataset} being released to facilitate research in this area. Additionally, these datasets fail to evaluate the consistency and reliability of model outputs, which in turn limits the performance of current systems.

In summary, the integration of LLMs with MVQA systems represents a significant advancement in intelligent medical research. These models offer the potential to improve diagnostic accuracy, enhance clinical decision-making, and provide valuable insights to healthcare professionals. Further research is needed to address insufficient data diversity and lack of consistency evaluation to realize the potential of these technologies.  

\subsection{Consistency and Reliability of Medical Large Language Models}
Medical LLMs are increasingly used for tasks such as medical diagnosis, treatment recommendation, and patient interaction. However, ensuring their consistent performance, especially in high-stakes environments, remains a significant challenge.

In scenarios where models operate as black boxes—lacking access to internal mechanisms and relying solely on API-based interactions, as is the case with commercial models like GPT-4 used in multimodal tasks—identifying unreliable responses becomes even more challenging. Without visibility into the model's inner workings, developers and users cannot inspect intermediate representations, attention weights, or confidence scores that might indicate why a response is erroneous or misleading.

The evaluation of medical LLMs also plays a crucial role in ensuring their reliability. Traditional metrics like accuracy and F1-score, while informative, may not fully capture the model’s consistency and reliability in clinical settings. To address this, recent work has introduced methods focusing on neighborhood consistency, where the reliability of LLM responses is inferred by evaluating response consistency over semantically equivalent question rephrasings \cite{khan2024consistency}. This approach, first introduced in the context of vision-language models, defines a “neighborhood” as semantically equivalent variations of an original question, generated as rephrasings using a proxy visual question generation (VQG) model. Consistency is then measured by assessing the agreement of a model's responses across these rephrasings, with higher consistency indicating greater reliability.

Additionally, Johnson et al.  \cite{johnson2019mimic} proposed a multi-dimensional evaluation framework that includes measures of robustness, interpretability, and fairness. This framework, combined with neighborhood consistency, has been instrumental in identifying potential weaknesses in LLMs before their deployment in real-world applications. 

While evaluating consistency in responses is a valuable step toward assessing the reliability of medical LLMs, it is essential to recognize that consistency alone does not guarantee the correctness of the information provided. A model may consistently produce the same answer to semantically equivalent questions, yet this answer could be consistently incorrect. This limitation underscores the need for evaluation frameworks that not only assess response consistency but also rigorously verify the factual accuracy and clinical validity of the model's outputs \cite{CHEN2025}. Incorporating measures of correctness alongside consistency is crucial to ensure that medical LLMs provide reliable and clinically sound information, thereby safeguarding patient care and upholding medical standards.

\subsection{Medical Visual Question Answering Datasets}
The availability of high-quality datasets is crucial for training and evaluating VQA models, as these datasets provide the necessary context for models to learn from. Over the years, several datasets have been developed to advance research in this area, each contributing unique aspects to the field.
The VQA-RAD dataset \cite{lau2018dataset} is an early example of a shared VQA dataset, including a diverse set of radiology images along with corresponding questions and answers, covering topics such as abnormalities, device identification, and anatomical location. This dataset has been widely used as a benchmark in the field.

\begin{itemize}
\item The SLAKE dataset \cite{liu2021slake} was developed to address the need for a more semantically rich and varied dataset. SLAKE includes radiology, pathology and dermatology images, with a greater emphasis on the diversity of questions. The inclusion of more detailed annotations has made SLAKE useful for training models that can handle complex and nuanced queries.
\item The PathVQA dataset \cite{he2020pathvqa} focuses on pathology images, contains over 30,000 question-answer pairs, and emphasises educational and diagnostic questions, making it useful for training models that can assist in medical education and decision-making.
\item The VQA-Med dataset \cite{ben2019vqa}, includes a broader range of medical images, such as those from CT scans and MRIs. VQA-Med was designed to challenge models with more complex questions that require deeper understanding and reasoning capabilities.
\item The RadQA dataset \cite{soni2022radqa} is an example of multimodal data including text, images, and structured data. It is a collection of radiology reports paired with corresponding images and structured data. RadQA aims to facilitate the development of models that can integrate multiple data sources to provide more accurate and context-aware answers.
\end{itemize}

Despite substantial progress, there remain challenges in creating datasets that can fully capture the complexity of real-world medical scenarios. Many existing datasets are limited in scope, focusing on specific modalities or question types. To address these limitations, ongoing research is exploring the development of more comprehensive datasets that include a wider variety of medical images and question types, as well as annotations that provide deeper semantic understanding.

\begin{figure*}
\centering
\includegraphics[width=1\textwidth]{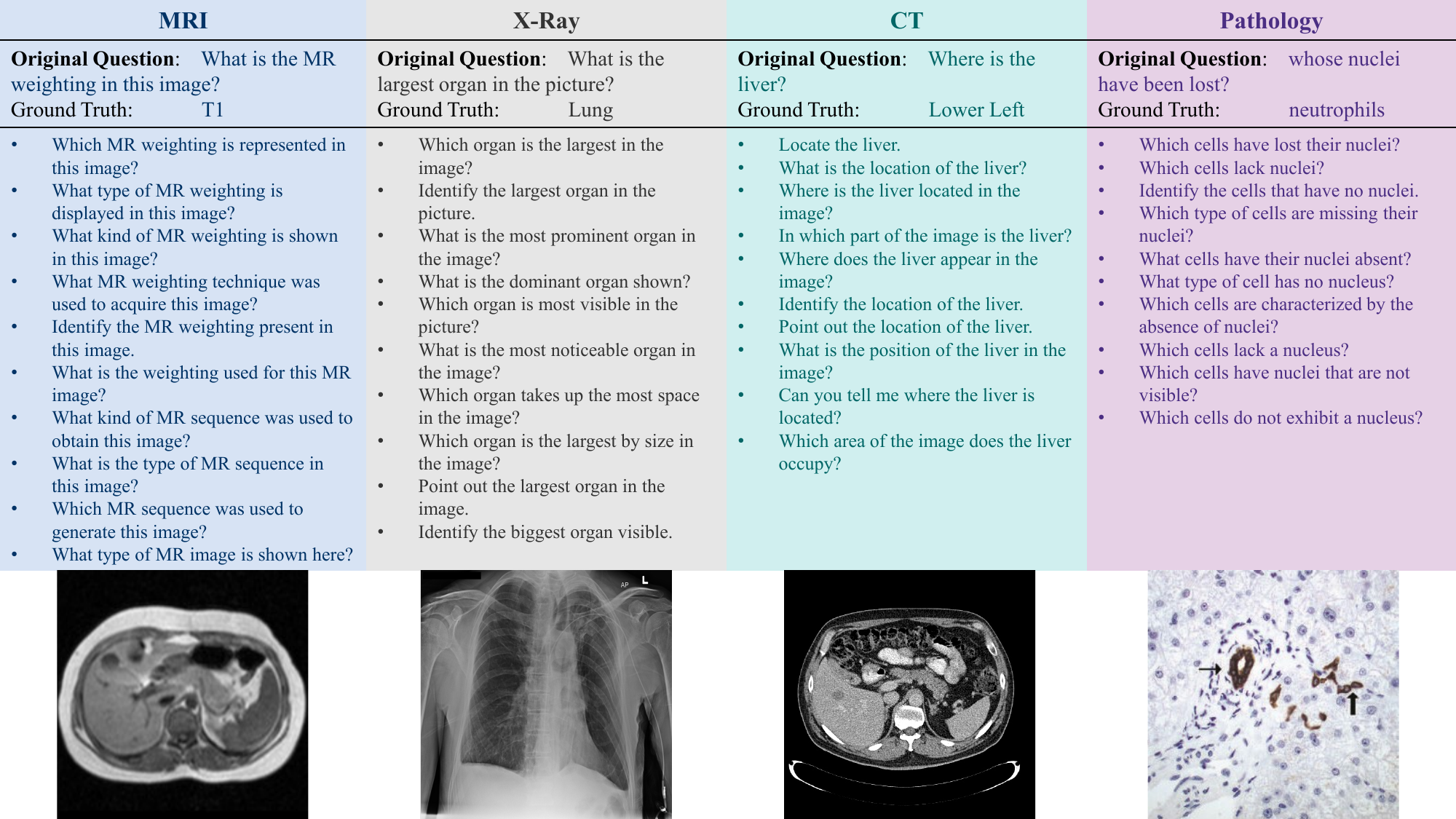}
\caption{Examples of different modalities (columns) in MVQA. The first row is the modality type, followed by the question and the list of the generated questions that have the same semantic meaning but with linguistic diversity with variations in syntax, structure, or phrasing.}
\label{fig_2}
\end{figure*}

% \begin{table}
%   \caption{Model Accuracy on SLAKE and VQA-RAD Datasets.}
%   \centering
%   \begin{tabular}{lcc}
%   \midrule
%   \textbf{Dataset}            & {\textbf{SLAKE}}               & {\textbf{VQA-RAD}}\\
%   \midrule
%   \textbf{Model Name}         & \textbf{ACC}                   & \textbf{ACC}\\
%   \midrule
%   \textbf{M2I2}               & 81.2                           & 76.8\\
%   \textbf{MUMC}               & 84.9                           & -\\
%   \textbf{BiomedGPT}          & 86.1                           & 73.2\\
%   \bottomrule
%   \end{tabular}
%   \label{table_1}
% \end{table}

\section{Methodology and Framework}

\subsection{Semantically Equivalent Question Augmentation}
This section introduces the Semantically Equivalent Question Augmentation (SEQA) framework. Figure\ref{fig_1} shows the workflow of our VQA augmentation method. 

First, we compile the original dataset. Each item in the dataset is processed using a large language model (LLM). The LLM generates multiple new questions for each original question while keeping the meaning unchanged. These generated questions vary in grammar (sentence structure or arrangement), structure (organizational framework), or phrasing (expression style). The prompt used for generation is as follows:

\begin{quote}
    The original question for the image is "\textit{question text}", and the original answer is "\textit{answer text}". Please generate 10 new questions with answers that have exactly the same meaning as the original question and answer (segment with a semicolon). Do not change the answer. The question needs to be kept in conjunction with the image I provided you. Do not add additional information to the question. It is necessary to ensure that newly generated questions are semantically equivalent to the original question. Just return new questions.
\end{quote}

Although the LLM operates as a black box, previous research suggests that the LLM-generated rephrased sentences are mapped closely to the original question in the feature space~  \cite{khan2024consistency}. This means that the rephrased questions are similar to nearby samples of the original question. As a result, these newly generated questions remain semantically equivalent to the original ones.

Figure\ref{fig_2} presents examples of medical question variations generated by the LLM using this prompt under different modalities, along with the ground truth answers and images. Since the ground truth answer remains the same for all variations, only the original question’s ground truth is marked in the figure. We observe that each generated question conveys the same meaning as the original.

These rephrased questions, along with their answers and corresponding images, are merged back into the dataset. The augmented dataset is then fed into the VLM. If the model provides consistent answers to semantically equivalent questions, it is considered a consistent model.

\subsection{Evaluation Metrics}
We proposed three evaluation metrics to assess the richness of the datasets: Average Number of QA Items per Image (ANQI), Average Number of Questions per Image with the Same Answer (ANQA), and Average Number of Open-Ended Questions per Image with the Same Semantics (ANQS).

\begin{itemize}
    \item ANQI measures the average number of QA pairs per image, reflecting the dataset's question density (Eq.\ref{eq_1}). A higher ANQI suggests that the dataset captures a broader range of visual-linguistic associations, encouraging models to generalize across varied QA contexts.
    \item ANQA measures the proportion of images where multiple questions receive the same answer, indicating answer consistency (Eq.\ref{eq_2}). ANQA captures answer redundancy. If a model can provide consistent answers to multiple questions yielding the same ground truth, it is likely to be more reliable in practical deployment. 
    \item ANQS measures the proportion of images where different open-ended questions with the same answer exist, capturing the presence of semantically equivalent questions in the dataset (Eq.\ref{eq_3}). Datasets with higher ANQS values better reflect natural language variability and thus offer a more realistic testing environment for model reasoning and generalization.
\end{itemize}

\begin{equation}
    \label{eq_1}
    ANQI = \frac{\text{Number of QA items}}{\text{Number of Images}}
\end{equation}

\begin{equation}
    \label{eq_2}
    ANQA = \frac{\text{Number of Images with same answers}}{\text{Number of Images}}
\end{equation}

\begin{equation}
    \label{eq_3}
    ANQS = \frac{\text{Number of open-end questions with same image and answer}}{\text{Number of Images}}
\end{equation}

We introduced accuracy, consistency level and, TAR-SC to evaluate model performance. The definition of consistency level and TAR-SC are as follows:
\begin{equation}
    \textit{Consistency Level}(n) = \textit{Maximum identical predictions for semantically equivalent variant questions}
\end{equation}
If a question generates ten variants and the model predicts answers for all, consistency is defined as the highest frequency of any single prediction. For example, if the ten predictions include three "1s" and four "2s," the consistency level is 4.

% \begin{equation}
%     \textit{Rejection Rate} = 
%     \frac{\textit{Total predictions occurring less than } n \textit{ times}}
%     {\textit{Total number of variant questions}}
%     \quad (\textit{Consistency Level} \geq n)
% \end{equation}
% For example, if n=2, the model’s predictions for the ten variant questions are A0, A0, A0, A1, A1, A2, A3, A4, A5, and A6. In this case, We only accept answers that appear at least twice in the predictions, designated as A0 and A1. Other responses, such as A2, A3, A4, A5, and A6, are rejected as they do not have another matching prediction. The rejection rate for the question is 50\%.

% \begin{equation}
%     Acc = 
%     \frac{\textit{Number of correct predictions among accepted answers.}}
%     {\textit{Number of accepted answers.}}
%     \quad (\textit{Consistency Level} \geq n)
% \end{equation}
% Under the same assumption, if the ground truth is A0, among the five accepted answers, three were correct, resulting in an accuracy of 60\%. We use accuracy to assess the model's reliability.

Previous research \cite{atil2024llm} introduced TAR to evaluate the consistency of model responses. Building on this, we propose an improved version that considers not only consistency but also accuracy. Total Agreement Rate with Similar Input and Correct Answer (TAR-SC) is a novel metric that calculates the mean score of correct answers for each set of semantically similar questions. TAR-SC assesses the model’s consistency in providing the same correct answer to different questions that have the same meaning.

\begin{equation}
    \textit{TAR-SC} = \textit{Mean} \left( \textit{Acc for each original questions} \right)
\end{equation}

Assuming that two original questions, \( Q_0 \) and \( Q_1 \), have variant questions \( Q_{00} \), \( Q_{01} \), \( Q_{02} \), \( Q_{03} \), \( Q_{04} \) and \( Q_{10} \), \( Q_{11} \), \( Q_{12} \), \( Q_{13} \), with corresponding answers \( A_{00} \), \( A_{00} \), \( A_{00} \), \( A_{01} \), \( A_{02} \) and \( A_{10} \), \( A_{10} \), \( A_{11} \), \( A_{12} \). The ground truth answers to these questions are \( A_{00} \) and \( A_{10} \), respectively. The accuracy for the \( Q_0 \) set would be 60\%, and the accuracy for the \( Q_1 \) set would be 50\%. Therefore, the TAR-SC would be 55\%.

By calculating the mean accuracy within each semantically equivalent question set, TAR-SC reflects the model’s ability to maintain both semantic understanding and factual correctness. It penalizes models that produce consistent yet incorrect responses, thus offering a more meaningful assessment of real-world consistency. In essence, TAR-SC bridges the gap between consistency-focused evaluation and accuracy-based assessment.

These metrics provide a comprehensive evaluation of both the accuracy and consistency of the models, which are crucial for applications in the medical domain where consistent information is essential.

\section{Experiments}
\subsection{Experimental Setup}
%To evaluate the consistency and correctness of MVQA models, w
We conducted comprehensive experiments using three state-of-the-art MVQA models: M2I2 (Masked image modeling, Masked language modeling, Image text matching and Image text alignment via contrastive learning) \cite{li2023self}, MUMC (Masked image and text modeling with Unimodal and Multimodal Contrastive losses) \cite{li2023masked}, and BiomedGPT  \cite{zhang2024generalist}. They employ distinct pretraining strategies and architectures, providing complementary assessments.
\begin{itemize}
\item \textbf{M2I2}: M2I2 is a multimodal MVQA model that integrates visual and textual information to answer clinical questions accurately \cite{li2023self}. M2I2 integrates masked modeling and contrastive learning for self-supervised vision-language pertaining. It employs attention mechanisms to align image regions with relevant question words, enhancing its ability to focus on pertinent image features when generating answers.
\item \textbf{MUMC}: The MUMC model leverages a unified framework to process medical images and textual data \cite{li2023masked}. MUMC introduces unimodal and multimodal contrastive losses to enhance alignment between image and text embeddings. It incorporates domain-specific knowledge graphs to improve reasoning and provides more accurate and contextually relevant answers by understanding complex medical concepts.
\item \textbf{BiomedGPT}: BiomedGPT is a generative pre-trained transformer model tailored for biomedical tasks \cite{zhang2024generalist}. BiomedGPT utilizes a unified encoder-decoder architecture to handle diverse biomedical tasks. It is capable of handling vision, language, and multimodal inputs, making it suitable for MVQA tasks that require understanding of both images and associated clinical questions.  
\end{itemize}
   
For the evaluation dataset, we augmented the SLAKE and VQA-RAD datasets using the SEQA framework with Gemini 1.5 Flash as the LLM. Compared to other LLMs, Gemini 1.5 Flash offers a cost-effective solution. This process resulted in new datasets referred to as SLAKE En 1.0-LD, VQA-RAD-LD and Path-VQA-LD,respectively. 

Next, we conducted experiments on SLAKE En 1.0-LD and VQA-RAD-LD. We chose to experiment only on SLAKE and VQA-RAD because these two datasets have been fully reviewed by human experts, ensuring higher reliability. Given the critical importance of annotation accuracy in medical image analysis, we prioritized datasets with comprehensive expert-validated annotations.

% PathVQA's annotations, though manually checked, originate from a semi-automatic extraction process. This method may introduce inconsistencies or ambiguities, potentially affecting the dataset's reliability for certain applications.

These 3 models then independently performed predictions on SLAKE En 1.0- LD and VQA-RAD-LD containing augmented questions to evaluate their consistency under two different scenarios.

We conducted experiments under two scenarios:

\begin{itemize}
\item Zero-shot: Models were tested without any task-specific fine-tuning on the augmented datasets. This scenario assesses the models’ inherent ability to generalize to new data and handle the linguistic variations introduced by the semantically equivalent questions. This demonstrates the limitations of these three MVQA models in providing consistent answers.
\item Fine-tuning: Models were fine-tuned on the training sets of the augmented datasets before evaluation. This process aims to adapt the models to the specific characteristics of the datasets, potentially improving their performance and consistency.
\end{itemize}

Text data was tokenized using model-specific tokenizers, ensuring compatibility with each model’s architecture.

\subsection{Datasets}
In this work, we utilized two key datasets: SLAKE and VQA-RAD. 

The SLAKE dataset  \cite{liu2021slake} consists of 642 medical images and 7,479 question-answer pairs, which cover 2 languages (Chinese and English), but we use only the English version in our work. The VQA-RAD dataset  \cite{lau2018dataset} consists of 315 radiology images, specifically focusing on diagnostic tasks related to radiology, with 3,515 associated question-answer pairs. 

% Additionally, we have summarized the accuracy of the three models on the original VQA-RAD and SLAKE datasets in Table \ref{table_1}.

\subsection{Motivating Experiments}

To evaluate the consistency of existing advanced MVQA models, we test their ability to understand deep semantic meanings. The results reveal inconsistencies in their responses.

\begin{figure*}
\centering
\includegraphics[width=1\textwidth]{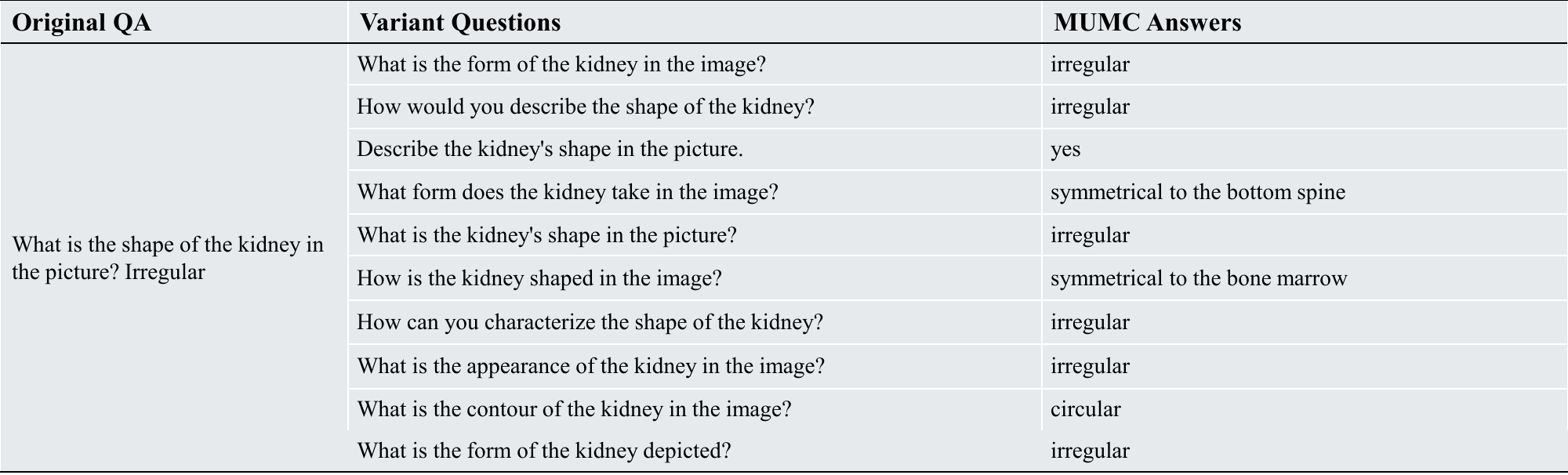}
\caption{Example of Semantically Equivalent Questions and Model Responses. "Original QA" refers to the original question and ground truth in the SLAKE dataset. "Variant Questions" are rephrased versions of the original question, designed to be semantically equivalent. The answers provided are generated by the MUMC model.}
\label{fig_3}
\end{figure*}

Figure\ref{fig_3} presents an example from the SLAKE dataset, showing the original question, its semantically equivalent variations, and the answers given by the MUMC model. We observe that when the same question is asked in different ways about the same image, the model often provides different answers. This indicates that the model's responses are not always reliable, even when the meaning of the questions remains unchanged.

%Additionally, Table\ref{table_3} shows the accuracy differences of M2I2, MUMC, and BiomedGPT on the original SLAKE and VQA-RAD datasets, as well as on the SLAKE En 1.0-LD and VQA-RAD-LD datasets, which include semantically equivalent questions. The results show large accuracy gaps across datasets. This suggests that none of these models can provide consistent answers for samples within the neighborhood of an original question.

\section{Results and Analysis}

\subsection{Experimental Results}

\begin{table}
  \caption{Comparison of Key Metrics Across Datasets. The best-performing value for each metric is highlighted in bold.}
  \centering
  \begin{tabular}{lllllll}
  \toprule
  \textbf{Dataset}  & \textbf{\# Modalities} & \textbf{\# Images} & \textbf{\# QA Items} & \textbf{\# ANQI} & \textbf{\# ANQA} & \textbf{\# ANQS} \\
  \midrule
  VQA-RAD         & 3          & 315              & 3,515     & 11.16           & 8.13            & 3.43         \\
  SLAKE-En 1.0    & 3          & 642              & 7,033     & 10.95           & 3.91            & 1.93         \\
  Path-VQA        & 2          & 4,998            & 32,799    & 6.56            & 6.27            & 1.55         \\
  OmniMedVQA      &\textbf{12} &\textbf{118,010}  & 127,995   & 1.08            & 1               & 1            \\
  VQA-RAD-LD      & 3	         & 314	            & 34,643	& \textbf{110.33}          & \textbf{106.03}          & 46.22        \\
  SLAKE En 1.0-LD & 3	         & 625	            & 68,451	& 109.52	      & 101.47	        & \textbf{66.76}        \\
  Path-VQA-LD     & 2	         & 4,553	        & 305,420	& 67.08	          & 66.11	        & 32.09        \\
  3 LDs merged    & 4          & 5,492    &\textbf{408,514}   &74.38   &73.41   &37.77\\
  \bottomrule
  \end{tabular}
  \label{table_2}
\end{table}

In Table \ref{table_2}, we compare the modalities, number of images, number of QA pairs, and ANQI, ANQA, and ANQS values of several commonly used MVQA datasets and our proposed augmented versions. The `LD' represents linguistically diverse datasets.

Table \ref{table_2} demonstrates that the augmented datasets are significantly more comprehensive than the original datasets. This is particularly evident in the three key metrics, where the generated datasets show a substantial increase compared to the original ones.

In ANQI, our dataset achieves an ANQI of 74.38, which is substantially higher than VQA-RAD (11.16), SLAKE (10.95), and PathVQA (6.56). A higher ANQI indicates that each image is associated with a larger number of question-answer pairs, enhancing the depth and richness of the dataset. This abundance allows models to learn more nuanced associations between visual features and corresponding questions \cite{shorten2019survey}.

In ANQA, we attain an ANQA of 73.41, compared to VQA-RAD’s 8.13, SLAKE’s 3.91, and PathVQA’s 6.27. A higher ANQA means that there are numerous differently phrased questions leading to the same answer for each image. This is crucial for training models to recognize that varied linguistic expressions can correspond to the same underlying concept, thereby improving the model’s ability to handle paraphrasing and synonymy \cite{wang2016deep}.

In ANQS, our dataset boasts an ANQS of 37.77, significantly surpassing VQA-RAD (3.43), SLAKE (1.93), and PathVQA (1.55). A higher ANQS reflects a greater variety of semantically equivalent open-ended questions for each image. This diversity exposes models to a wide range of semantic variations, enhancing their ability to generalize and understand nuanced language \cite{lake2017building}.     

Table \ref{table_3} summarizes the performance of the three models on the SLAKE En 1.0- LD and VQA-RAD-LD datasets under both zero-shot and fine-tuning scenarios as well as in-domain evaluation. In-domain evaluation refers to testing models on the original datasets it was trained on. The results in the "In-domain evaluation" are taken from the paper introducing the models. 

\begin{table}[htbp]
  \caption{Model performance comparison on different datasets in both zero-shot and fine-tuning scenarios. TAR-SC refers to Total agreement rate with the similar input and correct answer. ACC means accuracy. In-domain evaluation refers to testing models on the original datasets it was trained on. }
  
  \label{table_3}
  \centering
  \scriptsize % 修改字体大小
  \setlength{\tabcolsep}{4pt} 
  \begin{tabularx}{\textwidth}{l*{10}{X}} 
    & \multicolumn{2}{c}{\textbf{In-domain evaluation}} & \multicolumn{4}{c}{\textbf{Zero-shot}} & \multicolumn{4}{c}{\textbf{Fine-tune}} \\
    \cmidrule(lr){2-3} \cmidrule(lr){4-7} \cmidrule(lr){8-11}
    \textbf{Dataset} & \textbf{SLAKE En 1.0} & \textbf{VQA-RAD} & \multicolumn{2}{c}{\textbf{SLAKE En 1.0-LD (\%)}} & \multicolumn{2}{c}{\textbf{VQA-RAD-LD (\%)}} & \multicolumn{2}{c}{\textbf{SLAKE En 1.0-LD (\%)}} & \multicolumn{2}{c}{\textbf{VQA-RAD-LD (\%)}} \\
    \textbf{Metrics} & \textbf{ACC} & \textbf{ACC} & \textbf{ACC} & \textbf{TAR-SC} & \textbf{ACC} & \textbf{TAR-SC} & \textbf{ACC} & \textbf{TAR-SC} & \textbf{ACC} & \textbf{TAR-SC} \\
    \midrule
    \textbf{M2I2} & 81.2 & 76.8 & 64.39 & 56.76 & 62.34 & 63.65 & 77.15 & 69.44 & 67.28 & 69.55 \\
    \textbf{MUMC} & 84.9 & - & 59.77 & \textbf{59.3} & 68.37 & \textbf{68.32} & 76.14 & \textbf{72.4} & 74.78 & \textbf{74.68} \\
    \textbf{BiomedGPT} & 86.1 & 73.2 & 18.15 & 46.09 & 25.71 & 38.12 & 69.36 & 58.07 & 50.12 & 57.73 \\
    \bottomrule
  \end{tabularx}
\end{table}

% \begin{table}[htbp]
%   \caption{Model performance comparison on different datasets in both zero-shot and fine-tuning scenarios. Total agreement rate with the similar input and correct answer (TAR-SC), which is the mean score of correct answers for each set of similar questions.}
%   \label{table_3}
%   \centering
%   \small 
%   \setlength{\tabcolsep}{4pt} 
%   \begin{tabularx}{\textwidth}{l*{10}{X}} 
%     & \multicolumn{2}{c}{\textbf{In-domain evaluation}} & \multicolumn{4}{c}{\textbf{Zero-shot}} & \multicolumn{4}{c}{\textbf{Fine-tune}} \\
%     \cmidrule(lr){2-3} \cmidrule(lr){4-7} \cmidrule(lr){8-11}
%     \textbf{Dataset} & \textbf{SLAKE En 1.0} & \textbf{VQA-RAD} & \multicolumn{2}{c}{\textbf{SLAKE En 1.0-LD (\%)}} & \multicolumn{2}{c}{\textbf{VQA-RAD-LD (\%)}} & \multicolumn{2}{c}{\textbf{SLAKE En 1.0-LD (\%)}} & \multicolumn{2}{c}{\textbf{VQA-RAD-LD (\%)}} \\
%     \textbf{Metrics} & \textbf{ACC} & \textbf{ACC} & \textbf{ACC} & \textbf{TAR-SC} & \textbf{ACC} & \textbf{TAR-SC} & \textbf{ACC} & \textbf{TAR-SC} & \textbf{ACC} & \textbf{TAR-SC} \\
%     \midrule
%     \textbf{M2I2} & 81.2 & 76.8 & 64.39 & 56.76 & 62.34 & 63.65 & 77.15 & 69.44 & 67.28 & 69.55 \\
%     \textbf{MUMC} & 84.9 & - & 59.77 & 59.3 & 68.37 & 68.32 & 76.14 & 72.4 & 74.78 & 74.68 \\
%     \textbf{BiomedGPT} & 86.1 & 73.2 & 18.15 & 46.09 & 25.71 & 38.12 & 69.36 & 58.07 & 50.12 & 57.73 \\
%     \bottomrule
%   \end{tabularx}
% \end{table}

In Table \ref{table_3}, it can be observed that each model is influenced by the syntax, structure, or phrasing of the questions, with BiomedGPT being the most affected. After adding semantically equivalent questions, BiomedGPT’s accuracy dropped by 67.95\% on SLAKE and 47.49\% on VQA-RAD.

In the zero-shot setting, where models were evaluated without task-specific fine-tuning. M2I2 achieved an accuracy of 64.39\% on SLAKE En 1.0- LD and 62.34\% on VQA-RAD-LD. The TAR-SC scores were 56.76\% and 63.65\%, respectively. MUMC obtained an accuracy of 59.77\% on SLAKE En 1.0- LD and outperformed the other models on VQA-RAD-LD with an accuracy of 68.37\%. The TAR-SC scores were 59.30\% and 68.32\%, respectively. BiomedGPT showed lower performance in the zero-shot scenario, with accuracies of 18.15\% on SLAKE En 1.0- LD and 25.71\% on VQA-RAD-LD. The TAR-SC scores were 46.09\% and 38.12\%, respectively. 

After fine-tuning the models on the augmented datasets, M2I2 demonstrated significant improvement, achieving an accuracy of 77.15\% on SLAKE En 1.0- LD and 67.28\% on VQA-RAD-LD. The TAR-SC scores increased to 69.44\% and 69.55\%, respectively. MUMC continued to perform strongly, with accuracies of 76.14\% on SLAKE En 1.0- LD and 74.78\% on VQA-RAD-LD. The TAR-SC scores also improved to 72.40\% and 74.68\%, respectively. BiomedGPT exhibited the most significant improvement post fine-tuning, with accuracies rising to 69.36\% on SLAKE En 1.0- LD and 50.12\% on VQA-RAD-LD. The TAR-SC scores also increased substantially to 58.07\% and 57.73\%, respectively. 

\subsection{Performance Across Models}
The experimental results demonstrate disparities in model performance across in-domain, zero-shot, and fine-tuned scenarios. While all models achieve comparable in-domain accuracy on original benchmarks, BiomedGPT exhibits severe degradation under zero-shot testing on augmented datasets, revealing its vulnerability to semantic variations and rephrased questions. In contrast, M2I2 and MUMC maintain robust generalization, but still declined to a certain extent. In both zero-shot and fine-tuning scenarios, M2I2 excelled on SLAKE En 1.0-LD and MUMC dominated VQA-RAD-LD. Notably, MUMC achieves the smallest ACC-TAR-SC gaps across scenarios, indicating superior consistency in handling syntactic and structural variations, whereas BiomedGPT’s larger discrepancies highlight inconsistent outputs despite partial correctness. These findings underscore BiomedGPT’s architectural limitations in processing paraphrased inputs, contrasting with M2I2’s scenario-specific optimization and MUMC’s balanced performance, positioning the latter as the most clinically viable due to its dual emphasis on accuracy and semantic stability. 

%Analyzing the models’ performance across the two datasets reveals several interesting observations. SLAKE En 1.0-LD, with its diverse image modalities and complex questions, posed significant challenges for the models, reflected in variations in accuracy across datasets. The performance of MUMC reflects its architecture and reliance on domain-specific knowledge graphs, which align well with the radiology-focused nature of VQA-RAD-LD. Fine-tuning demonstrated a clear impact on all models, with each model benefiting from the process in unique ways. BiomedGPT showed significant changes after fine-tuning, indicating the importance of adapting pre-trained models to specialized tasks for achieving effective performance.

% \subsection{Correlation Between Accuracy and Consistency}
% As shown in Figure \ref{fig_4}, a positive correlation between ACC and TAR-SC was observed across all models and scenarios. Models with higher accuracy consistently achieved higher TAR-SC scores, indicating that overall correctness is closely linked to consistency in handling semantically equivalent questions. Additionally, fine-tuning demonstrated a significant impact, not only improving accuracy but also enhancing the models' ability to provide consistent answers, as reflected in increased TAR-SC scores. This highlights the importance of both accuracy and consistency in developing robust MVQA models.

\subsection{Analysis of TAR-SC Scores}
The TAR-SC metric provides insights into the models' consistency by focusing on the accuracy of semantically similar questions rather than overall accuracy across the entire dataset. This makes it a more suitable measure for evaluating how well models understand the underlying semantics of a question and their ability to provide consistent and correct answers.

In the zero-shot setting, MUMC demonstrated the highest TAR-SC scores, particularly on VQA-RAD-LD. Across all models, significant improvements in TAR-SC were observed after fine-tuning. This indicates that exposure to semantically equivalent questions during training plays a critical role in enhancing model consistency. Among the three models, MUMC consistently achieved higher TAR-SC scores compared to M2I2 and BiomedGPT, suggesting that its architecture is particularly well-suited to capturing semantic nuances in medical questions.

\subsection{Impact of Dataset Augmentation}

The augmentation of the SLAKE  \cite{liu2021slake} and VQA-RAD  \cite{lau2018dataset} datasets resulted in a substantial increase in the average number of question-answer pairs per image. 

The augmented datasets, SLAKE En 1.0- LD and VQA-RAD-LD, provided a more challenging and comprehensive evaluation for the models. The increase in the Average Number of Questions per Image with the Same Answer (ANQA) and the Average Number of Open-Ended Questions per Image with the Same Semantics (ANQS) indicates that models trained on these datasets are exposed to more linguistic diversity enhancing their ability to generalize and understand nuanced language  \cite{lake2017building}.

The augmentation of datasets with semantically equivalent questions had a positive impact on model performance. While fine-tuning led to improvements in both accuracy and TAR-SC across all models, the overall accracy remained lower than that on the original datasets. This indicates that the increased linguistic diversity posed new challenges for the models, as it tested their natural language understanding capabilities and their ability to recognize that different phrasings can express the same meaning. Fine-tuning on these augmented datasets enabled the models to learn from a broader range of question formulations, leading to improvements in both accuracy and consistency. By simulating such variability, the augmented datasets contribute to making models more adaptable in practical applications \cite{zhang2023large}.

\section{Discussion}
% The primary objective of this study was to enhance the consistency and reliability of MVQA models by augmenting existing datasets with semantically equivalent questions. Our experimental results demonstrate that this augmentation significantly improves the performance of state-of-the-art models, both in terms of accuracy and consistency, as measured by the Total Agreement Rate with Similar Input and Correct Answer (TAR-SC).

% \subsection{Impact of Dataset Augmentation}
% The augmentation of the SLAKE \cite{liu2021slake} and VQA-RAD \cite{lau2018dataset} datasets resulted in a substantial increase in the average number of question-answer pairs per image. 

% The augmented datasets, SLAKE En 1.0- LD and VQA-RAD-LD, provided a more challenging and comprehensive evaluation for the models. The increase in the Average Number of Questions per Image with the Same Answer (ANQA) and the Average Number of Open-Ended Questions per Image with the Same Semantics (ANQS) indicates that models trained on these datasets are exposed to more linguistic diversity enhancing their ability to generalize and understand nuanced language.

\subsection{Model Consistency}
\begin{figure*}
\centering
\subfigure[M2I2 SLAKE En 1.0- LD]{\includegraphics[width=0.497\textwidth]{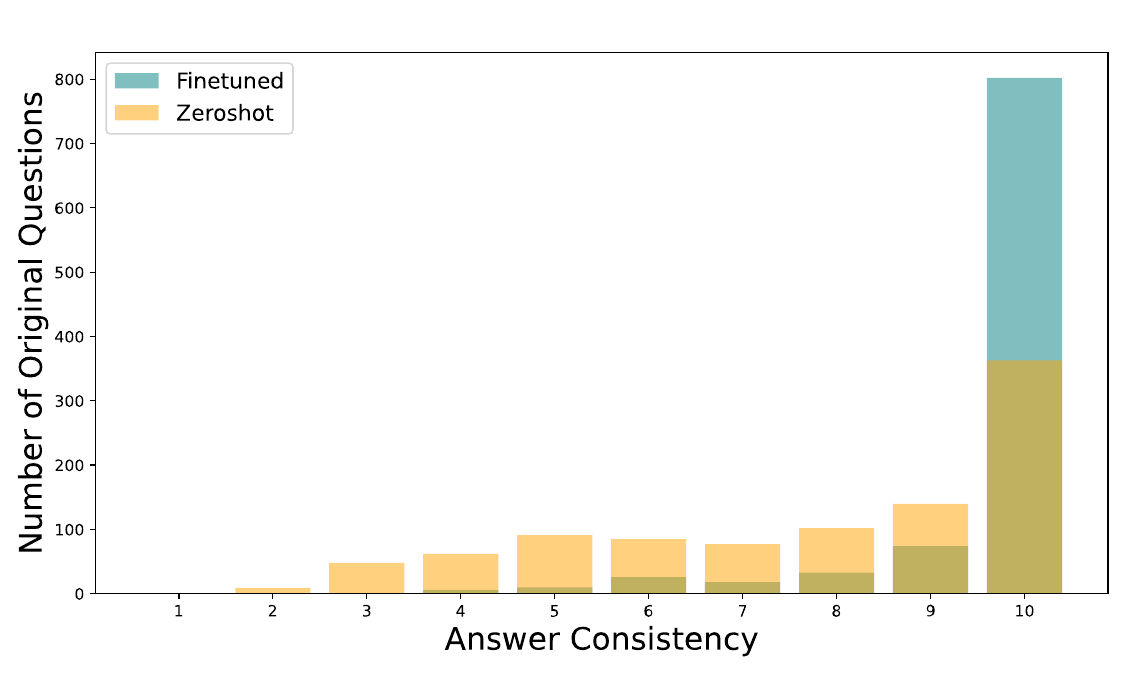}}
\subfigure[M2I2 VQARAD-LD]{\includegraphics[width=0.497\textwidth]{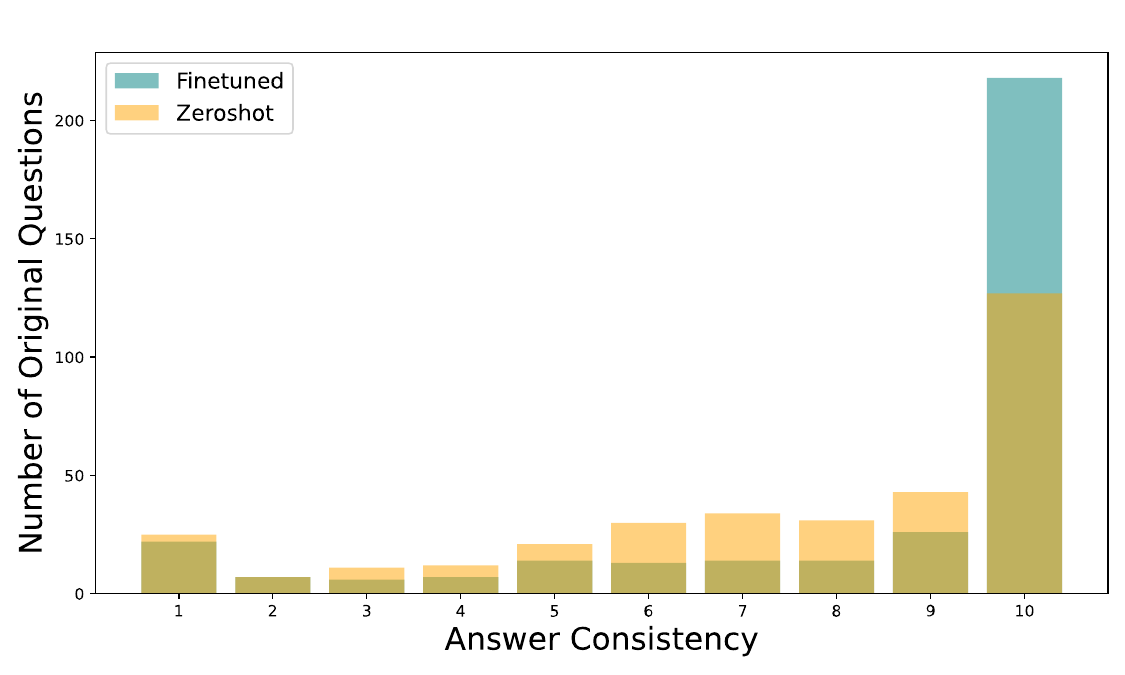}}

\subfigure[MUMC SLAKE En 1.0- LD]{\includegraphics[width=0.497\textwidth]{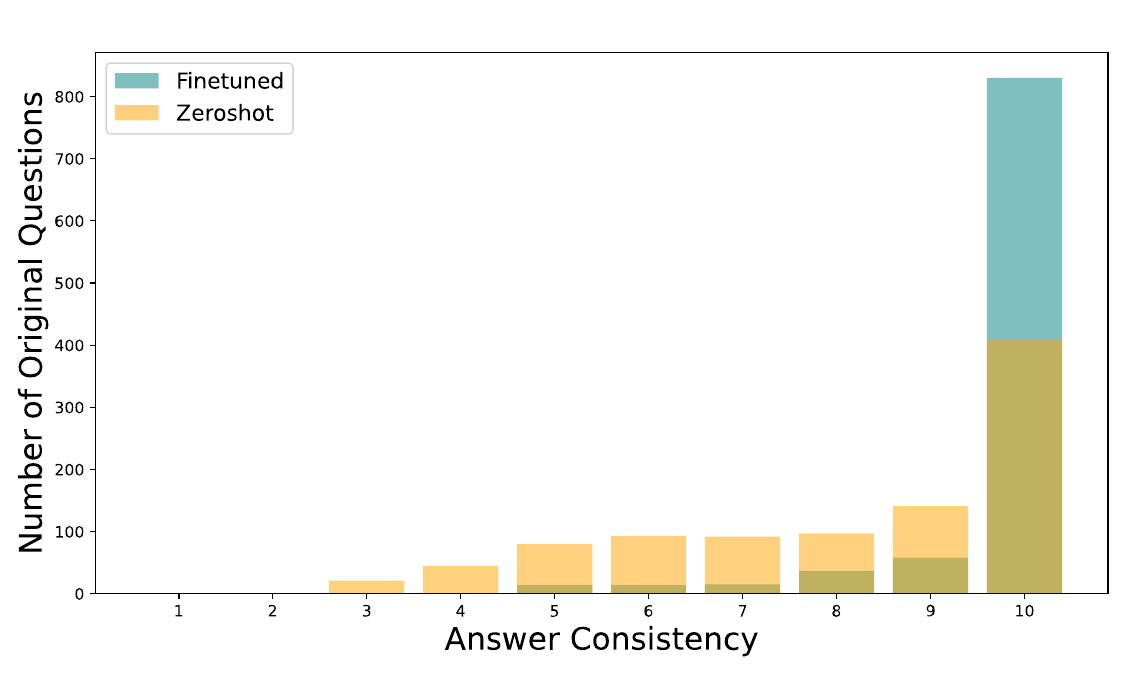}}
\subfigure[MUMC VQARAD-LD]{\includegraphics[width=0.497\textwidth]{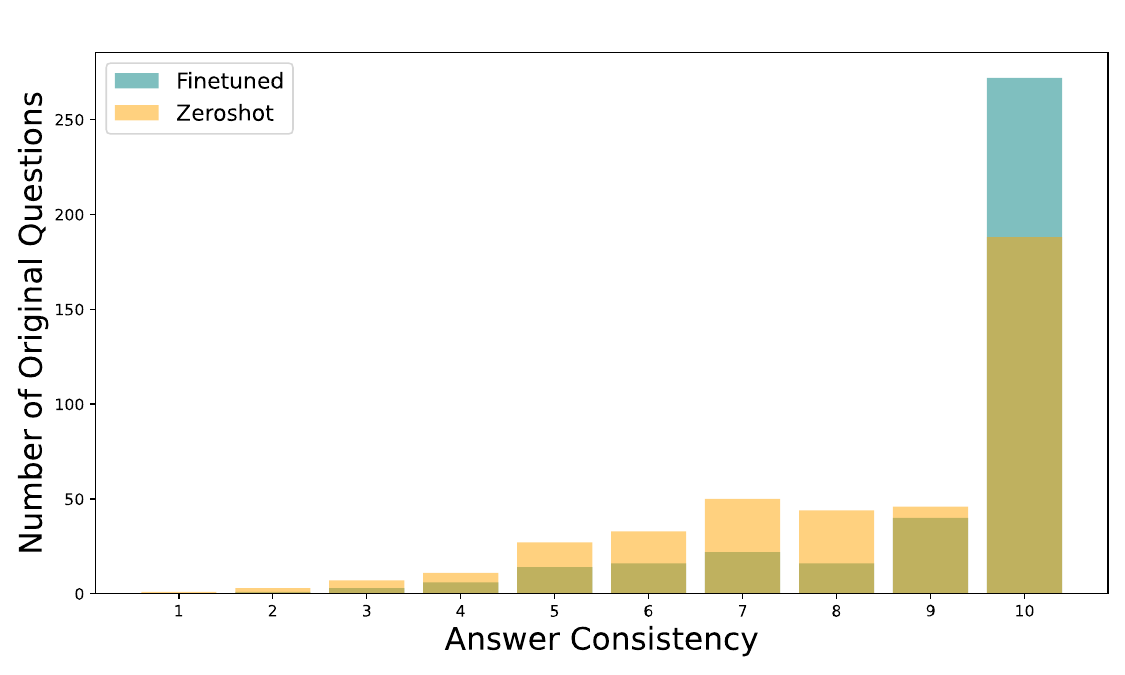}}

\subfigure[BiomedGPT SLAKE En 1.0- LD]{\includegraphics[width=0.497\textwidth]{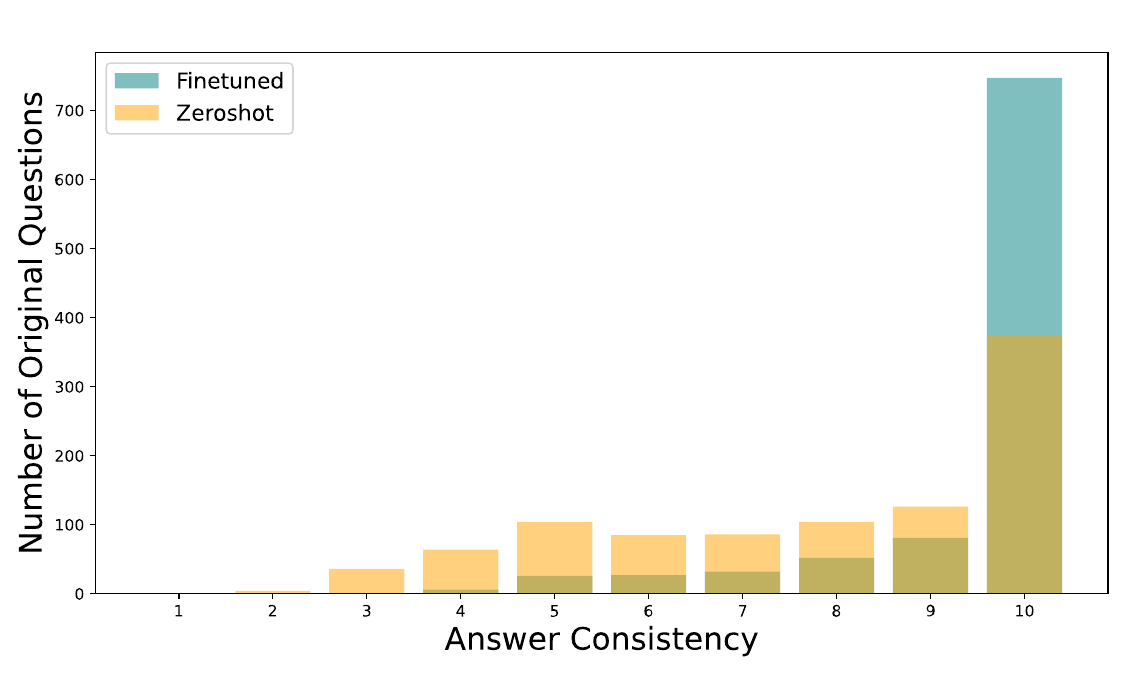}}
\subfigure[BiomedGPT VQARAD-LD]{\includegraphics[width=0.497\textwidth]{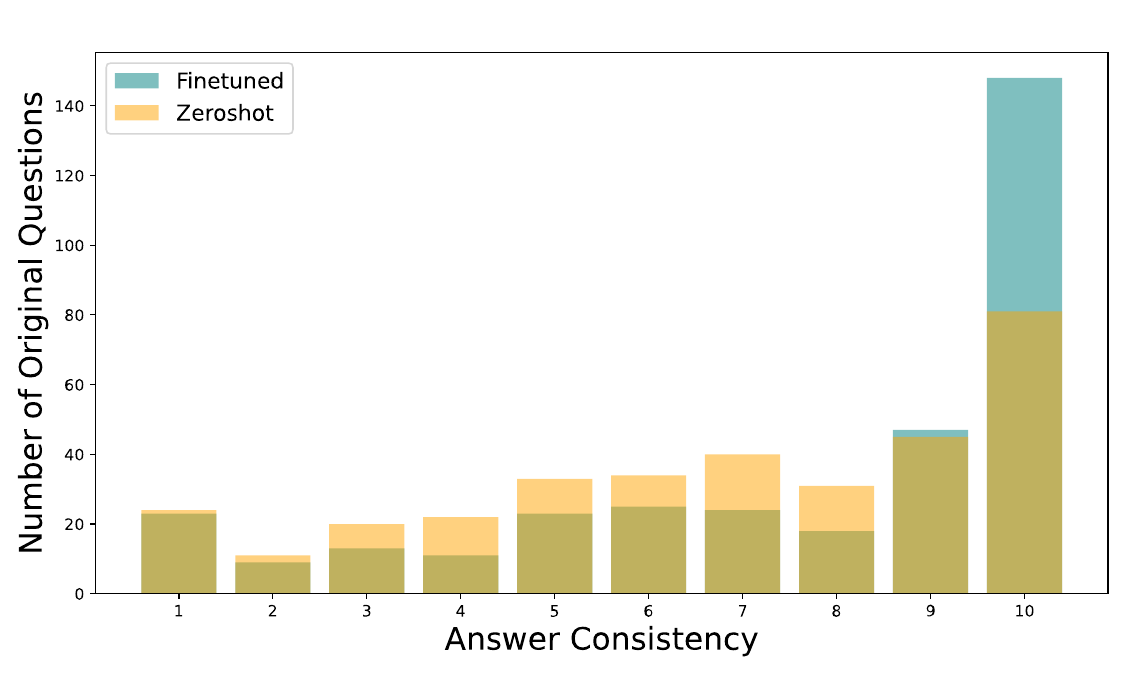}}

\caption{Distribution of Original Questions by Answer Consistency Levels. The x-axis represents answer consistency levels (the number of variation questions that provided the same answer), while the y-axis shows the number of original questions corresponding to each level.}
\label{fig_5}
\end{figure*}
After fine-tuning the MVQA model on our dataset, we observed changes in consistency. As shown in Figure \ref{fig_5}, all the models have exhibited a significant improvement in consistency. In Figure \ref{fig_5}, after fine-tuning, the answers to all ten variation questions for each model on each dataset tend to become consistent.

Our experiments revealed that all the models benefited from fine-tuning on the augmented datasets. Based on Figure \ref{fig_5}, it is evident that the augmented datasets offer increased diversity in question formulation, which enhances the consistency of the models’ responses. Furthermore, Table \ref{table_3} indicates that the three models showed improvements in both TAR-SC and accuracy (ACC) after fine-tuning. This demonstrates that fine-tuned models not only provide more accurate responses but also exhibit higher consistency, being less influenced by the syntax, structure, or phrasing of the questions. High accuracy is essential for correctness, but consistency ensures that models provide consistent answers regardless of variations in question phrasing. This is particularly critical in the medical domain, where inconsistent responses can lead to confusion or misinterpretation.

Notably, the MUMC model consistently outperformed the other models in terms of TAR-SC, particularly on the VQA-RAD-LD dataset. This suggests that MUMC’s architecture is adept at capturing semantic nuances and maintaining consistency in responses to semantically similar questions.

The M2I2 model also demonstrated strong performance, especially after fine-tuning, indicating its effectiveness in integrating multimodal information and adapting to the augmented datasets. BiomedGPT, while initially underperforming in the zero-shot scenario, showed significant improvements post fine-tuning. This underscores the importance of task-specific training for large generative models in specialized domains \cite{zhang2024generalist}.

\subsection{Comparison with Reasoning Models}
While reasoning models have gained prominence for their enhanced capabilities in complex problem-solving and decision-making tasks, they are not immune to consistency issues. These models predominantly rely on recognizing patterns and statistical correlations within extensive datasets, rather than truly comprehending underlying concepts \cite{toloka2024reasoning}. This fundamental limitation can lead to inconsistencies in their outputs, especially when faced with novel or ambiguous inputs.

For instance, large language models (LLMs) like GPT-4 have demonstrated impressive performance across various domains. However, studies have shown that these models can produce contradictory responses to semantically equivalent prompts, indicating a lack of genuine understanding and consistent reasoning \cite{ahn2025prompt}. This phenomenon is partly due to their dependence on surface-level patterns in data, which may not capture the deeper semantic relationships necessary for consistent reasoning.

Moreover, the absence of a true causal understanding further exacerbates these consistency issues. LLMs can identify statistical correlations but often fail to grasp causal relationships, leading to outputs that may be contextually appropriate but logically flawed \cite{white2025llms}. This limitation underscores the challenge of achieving consistency, as the models may not reliably discern the implications of their generated content.

\subsection{Limitations and Future Work}
This study has several limitations that highlight opportunities for future work. One of the primary limitations of our approach is its reliance on the performance of LLMs for generating semantically equivalent questions. During our experiments, we encountered instances where the LLMs failed to produce the desired number of question variants. This inconsistency suggests that the LLMs may introduce errors during the generation process. If these errors are not carefully addressed, they could propagate into the augmented dataset, potentially compromising its quality and, in turn, affecting the performance of the trained models  \cite{bender2021dangers}. 

Another limitation is the relatively narrow scope of the datasets used in this study. While SLAKE and VQA-RAD are well-established benchmarks, their limited size and diversity may restrict the generalizability of our findings to broader clinical settings. Incorporating larger and more diverse datasets, such as OmniMedVQA \cite{hu2024omnimedvqa} and PathVQA \cite{he2020pathvqa}, which includes a wider range of medical imaging modalities, could help address this issue and improve the robustness of the models.

Finally, while the evaluation metrics employed in this study provide an alternative overview of model performance, they may not fully capture certain nuances relevant to real-world clinical applications. For instance, the ability to handle ambiguous or context-dependent questions, which is critical in medical practice, requires more sophisticated evaluation frameworks. Future work could explore the development of metrics tailored to these specific challenges.

Future work will focus on expanding the framework to test on larger and more diverse datasets— for example, by integrating datasets such as OmniMedVQA, PathVQA, and VQA-Med — could further enhance model robustness and generalizability. We will also evaluate our model in real-world clinical scenarios, incorporating feedback from medical professionals to refine their utility and safety in healthcare environments \cite{rasmy2021med}. Additionally, exploring non-medical applications of the framework, particularly in domains where consistency is critical, such as legal or educational question answering, will be explored for its ability to demonstrate its broader utility \cite{ji2023survey}. 

\section{Conclusions}
This study presented a novel approach to enhancing the consistency of MVQA models through dataset augmentation using semantically equivalent questions. Our experiments demonstrate that our augmentation improved consistency as measured by TAR-SC. The findings suggest that exposing models to a wider variety of question formulations during training enables them to generalize better and provide consistent answers, which we suggest is essential for their adoption in clinical practice. The success of models like MUMC and M2I2 in handling the augmented datasets indicates that deep linguistic understanding are well-suited for MVQA tasks.  The substantial improvements observed in BiomedGPT after fine-tuning highlight the potential of large generative models when appropriately adapted to specialized domains.

The enhanced consistency of MVQA models have significant implications for clinical practice. Models that can understand and respond accurately to varied question formulations are more likely to be trusted by healthcare professionals. They can serve as valuable tools for assisting in diagnosis, education, and patient communication \cite{esteva2019guide}.  

\section*{Acknowledgements}
I would like to express my sincere gratitude to Ms. Chunyu Liu for her insightful contributions to the conceptualization of evaluation metrics, and for her thoughtful assistance in refining the clarity and grammatical accuracy of the manuscript.

\bibliographystyle{unsrt}  
\bibliography{references}

\end{document}